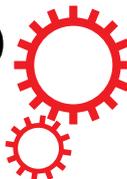

OPEN

# Rapid Bayesian optimisation for synthesis of short polymer fiber materials

Cheng Li[1], David Rubín de Celis Leal[2], Santu Rana[1], Sunil Gupta[1], Alessandra Sutti[2], Stewart Greenhill[1], Teo Slezak[2], Murray Height[3] & Svetha Venkatesh[1]

The discovery of processes for the synthesis of new materials involves many decisions about process design, operation, and material properties. Experimentation is crucial but as complexity increases, exploration of variables can become impractical using traditional combinatorial approaches. We describe an iterative method which uses machine learning to optimise process development, incorporating multiple qualitative and quantitative objectives. We demonstrate the method with a novel fluid processing platform for synthesis of short polymer fibers, and show how the synthesis process can be efficiently directed to achieve material and process objectives.

Experimentation is central to the discovery and development of processes for the synthesis of novel materials. This involves a myriad of decisions for the system configuration, operating parameters and material properties, described by multiple variables. As the multi-variable complexity of new processes and materials increases, the task of identifying the best combination of process settings to achieve output material property targets becomes complex and time consuming. Design of Experiment (DoE) methodologies[1–3] (including Response Surface Analysis) have for decades provided experimentalists with a framework to systematically explore multi-variable space, however the experimental costs can become burdensome as complexity increases.

Challenges posed by materials processing optimisation are manifold. Material and process variables must be combined to search for optimal target characteristics. This is complex since the variables relating to material properties and processes are often non-homogeneous or non-orthogonal. Further, multiple diverse objectives need to be incorporated; these could be properties evaluated through quantitative measures, or otherwise be specified qualitatively due to need or difficulty of attaining quantitative values - for example morphological homogeneity or presence of by-products. Cost, performance, and quality metrics are typical for industrial systems. Finally, while existing databases of materials and properties may provide varying degrees of guidance for synthesis of novel materials, experimenters in the early stages of process development and optimisation rely more upon DoE approaches, hypothesised governing principles, broad approximations and expert intuition.

In chemistry and materials science, computational methods such as Density Functional Theory (DFT) can be used to predict interactions at the atomic scale. In combination with such methods, machine learning models have been used to predict material properties, model inter-atomic potentials, predict crystal and lattice structures, and understand property-structure relationships in amorphous materials[4]. Using an iterative design process, different material compositions have been formulated and evaluated computationally in order to optimise particular physical properties, such as the elastic modulus of M2AX compounds[5], or thermal conductance across nano-structures[6]. Complex systems such as particle-reinforced composites have recently been approached through multistep modelling[7] which demonstrates efficient tuning of materials properties and composition through micro-mechanical and multiscale models.

Computational evaluations are usually easier to perform than physical experiments, allowing many more iterations to be done in the search process. In the absence of detailed knowledge of a material synthesis process, and in the absence of sensitive predictive tools like DFT, finite element analysis, and computational fluid-dynamics, the optimisation of a physical process is best approached as a black-box problem, to reduce the impact of possibly-incorrect or inaccurate assumptions. When materials must be physically synthesised, *efficient global optimisers* are preferred. These have been recently used to design new shape memory alloys from a large search space of compositions[8], and to design experiments for the production of Bose-Einstein Condensates[9].

[1]Centre for Pattern Recognition and Data Analytics (PRaDA), Deakin University, Victoria, Australia. [2]Institute for Frontier Materials (IFM), Deakin University, Victoria, Australia. [3]HeiQ Australia, Pty Ltd, Geelong, Victoria, Australia. Correspondence and requests for materials should be addressed to S.G. (email: s.greenhill@deakin.edu.au)

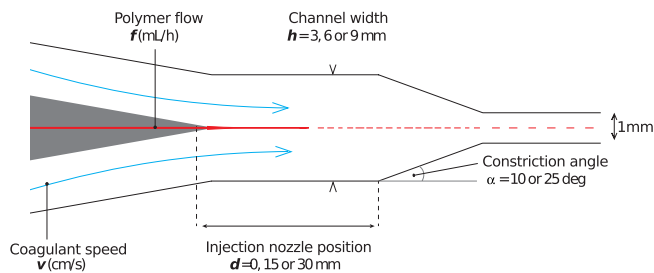

**Figure 1.** Diagram of short polymer fiber (SPF) synthesis using a microfluidic device and a polymer solution with higher viscosity than the coagulant. A primary breakup occurs as the polymer filament is elongated by the flowing coagulant during its gelation. A secondary breakup of the fibers can be induced by shear forces at the constriction of the laminar flow.

The computational challenge in optimisation of novel processes arises because the mathematical relationship between the control variables and the target is often unknown - it is a classic "Black-Box Function". Experimental data is expensive to acquire, so achieving an optimal setting with a minimum of experiments is desirable. Bayesian optimisation[10–12] is a machine learning framework to optimise expensive black-box functions, which it models through a stochastic Gaussian process (GP)[13]. The GP may be used to mimic an experimenter, building a mathematical model from available experimental data, then using its model it derives an "educated guess" to recommend the next experimental setting. The GP is updated as experiments proceed - the "educated guess" becomes increasingly accurate, and its *epistemic* uncertainty reduces. In contrast, there could be other sources of uncertainty, such as those stemming from uncertain inputs or from a stochastic system. Such uncertainties are *aleatoric* in nature and it is important to perform sensitivity analysis[14] of the result. Bayesian optimisation is theoretically shown to have the best achievable order of convergence rate in terms of number of samples (sub-linear growth in cumulative regret) for a global optimisation algorithm[15] and it allows integration of different output types such as numerical, pairwise quality measure and even their hybrid. It also offers high dimensional optimisation[16], optimisation with unknown constraints[17], and even transfer learning from past experiments[18] in an efficient optimisation framework. While Bayesian optimisation offers a powerful construct for adaptive experimentation[5], limited work exists[8,9] and does not tackle the key challenges of incorporating multiple objectives, or combining heterogeneous objectives.

In this contribution, we demonstrate that machine learning techniques can be used to effectively optimise developmental stage processes for synthesis of novel materials. We formulate a method based on Bayesian optimisation to help navigate experimental complexity by integrating an iterative experimental approach with data-driven models. Using both material and process-related variables as inputs, this Adaptive Experimental Optimisation (AEO) framework recommends the next experimental setting, followed by performance of the experiment and return of the data values to the algorithm in an iterative cycle until the target product goal is achieved. Multiple objectives that can be both quantitative and qualitative are accounted for in the framework.

We test the algorithm using a modular fluid processing platform to produce short (and ultrafine) polymer fibers[19,20], a novel physical process which has not previously been modelled or fully characterised. Short polymer fibers (SPF) are filamental micro structures with typical dimension magnitudes of 1–10 μm diameter and 10–1000 μm in length. SPF are produced by coagulating a polymer solution in a shear flow[21]. The fiber diameter and length are adjustable by varying the synthetic conditions, but little is known about the inter-parameter relationships in the system. In the new method described in this paper, a fiber is formed in a fluidic device by injecting a polymer solution coaxially into a rapidly flowing coagulant liquid (Fig. 1). The coagulant flows through a channel of defined width ($h$) and the polymer solution is injected at the centre line of the channel. The flow is then funnelled into a 1 mm wide output channel through a constriction of variable angle ($\alpha$). The distance between polymer injection and the start of the constriction ($d$) has a direct effect on the residence time of the polymer filament in the chamber before it is accelerated, and possibly elongated, within the constriction. The many degrees of freedom of this system, the dimensions (micron and submicron), the varied morphology of the products (from fibres, through to spheres and debris), and the complexity of the flow make modelling this system very challenging and of low accuracy. These features make this platform suitable for optimisation through AEO.

The main contributions of this paper are:

- Formulation of Adaptive Experimental Optimisation (AEO) incorporating both material properties and process characteristics within a coherent experimental optimisation framework;
- Incorporation of multiple objectives for experimental optimisation, including both qualitative and quantitative targets;
- Systematic methodology to objectively draw out insights about qualitative targets;
- Description of a novel process for synthesis of short polymer fibers (SPF);
- Validation of the AEO for optimisation of SPF materials and associated process.

## Methods

**Short Polymer Fiber Synthesis.** The fluidic devices were constructed with various combinations of the three geometric variables shown in Fig. 1, namely channel width ($h = 3$, 6 or 9 mm), injection position ($d = 0$, 15 or 30 mm) and constriction angle ($\alpha = 10°$ or $25°$), giving a total of 18 different configurations.

The coagulant liquid used was 1-butanol (>99%, Chem Supply [127]) maintained at a temperature of 4–10 °C and driven by a gear pump (Micropump GB-P23.KF5SA) connected to a variable power supply (TTiEX355) at a maximum current of 1 Amp. The voltage of the power supply was adjusted to control the coagulant flow rate. Poly(ethylene-co-acrylic acid) pellets (Primacor 5990I DOW [126]) were added to an aqueous solution of de-ionised water and ammonium hydroxide under reflux at 110 °C, to prepare a 16.5% w/v dispersion (polymer dope). This polymer dope was pumped into the device using 5 ml Terumo syringes mounted on a syringe pump (Legato 270). For each experiment the device was initially flushed with 50–100 ml of water through the coagulant inlet, then the polymer flow was initiated and once the polymer reached the nozzle (judged through observation of the flow through transparent tubing) the coagulant flow was started. After reaching a stable flow (typically 10 s), approximately 30 ml of produced suspension was collected in a glass beaker. Both polymer and coagulant flows were then stopped and the device flushed with water to avoid coagulation in the channels. The collected suspension was diluted in ethanol and a few drops were deposited on a glass slide and dried in air. Three values were explored for polymer flow rate (f = 80, 110, or 140 ml/h) and coagulant speed (v = 43, 68, 93 cm/s). In combination with geometric variables there are thus 162 possible experiments.

Microscopy images of the dry samples were taken at different magnifications using an optical microscope (Olympus BX51 with a DP71 camera) calibrated using a TEM 462 nm grid and an optical 10 μm calibration slide. ImageJ (Fiji distribution) was used to analyse the images for measurement of fiber dimensions. Fiber length was measured using the multi-line tool at 5x and 20x magnifications, while the single line tool was used for measurements of diameter at 100x magnification. High magnification and low field of view prevented the simultaneous measurement of length and diameter for each fibre, so separate length and diameter distributions were used. The median values of the measured length and diameter were extracted for each image. For length and diameter, a minimum of 50 or as many measurements as possible (if fewer fibers found) were taken per image, and a few images were processed per sample. The qualitative assessment used for the pairwise comparison ranking (see "Qualitative Score - $f_Q(\cdot)$ and interface for qualitative comparison") was performed with the objective to emphasise fiber suspensions with less by-product (debris and spheres) and less entangled fibers.

**Adaptive Experimental Optimisation.** The role of AEO is to guide the experimenter by suggesting the next experimental setting based on Bayesian optimisation with the objective to reach the target output properties. Targets are specified and the experimentally available range of both process and product parameters are stipulated. In our case, the product parameters (**y**) are median length, median diameter and quality of fibers; and the process parameters (**x**) are position, constriction angle, channel width, polymer flow and solvent flow.

Figure 2 describes the overall iterative optimisation process. Step 1 models an unknown function with a Gaussian process using the data acquired thus far. Step 2 recommends the next experimental setting ($\mathbf{x}_t$) by optimising an acquisition function derived from the Gaussian process. In Step 3 the experimentalist performs the suggested experiments and measures the output characteristics, as described in the Methods section. Step 4 uses these measurements to compute the combined target utility $y_t$ (Equation (10)). This new observation pair ($\mathbf{x}_t, y_t$) is appended to the data set. Step 5 checks whether the product meets the target specifications. Steps 1–5 are repeated in an iterative loop until the target is achieved or process is terminated. Steps 1, 2 and 4 are described in the following sections.

**Step 1: Function modelling.** Bayesian optimisation is a powerful tool to optimise an expensive black-box function and we use it to recommend the next experiment setting based on the observations $\{x_{1:t}, y_{1:t}\}$, where $y_{1:t}$ are the objective values computed via experimentation by evaluating the results. In our case a multi-objective function combines distance to target length and diameter and the qualitative score via Equation (10). A common method to model an unknown function $f$ is using a Gaussian process as a prior, that is, $f(\cdot) \sim GP(0, \mathbf{K}(\cdot, \cdot))$. The prior probability of function values $y_{1:t}$ is a multivariate Gaussian where

$$p(y_{1:t}) = |2\pi \mathbf{K}| \exp\left(-\frac{1}{2} y_{1:t}^T \mathbf{K}^{-1} y_{1:t}\right) \tag{1}$$

and **K** is the $t \times t$ covariance matrix whose $ij$-th element is $k(\mathbf{x}_i, \mathbf{x}_j)$. The $k$ is a kernel function. The kernel function computes the distance between data points. A common kernel function is the squared exponential (SE) function, which is defined as

$$k(\mathbf{x}_i, \mathbf{x}_j) = \exp\left(-\frac{1}{2\theta^2} ||\mathbf{x}_i - \mathbf{x}_j||^2\right) \tag{2}$$

where the kernel length-scale $\theta$ affects the smoothness of the objective function - a larger value means a smoother function.

When a new observation point $\mathbf{x}_{t+1}$ is made, the predictive distribution of $y_{t+1}$ can be computed as

$$y_{t+1}|y_{1:t} \sim \mathcal{N}(\mu_{t+1}(\mathbf{x}_{t+1}|\mathbf{x}_{1:t}, y_{1:t}), \sigma_{t+1}^2(\mathbf{x}_{t+1}|\mathbf{x}_{1:t}, y_{1:t})) \tag{3}$$

with $\mu_{t+1}(\cdot) = \mathbf{k}^T \mathbf{K}^{-1} y_{1:t}$ and $\sigma_{t+1}^2(\cdot) = k(\mathbf{x}_{t+1}, \mathbf{x}_{t+1}) - \mathbf{k}^T \mathbf{K}^{-1} \mathbf{k}$, where $\mathbf{k} = [k(\mathbf{x}_{t+1}, \mathbf{x}_1) \cdots k(\mathbf{x}_{t+1}, \mathbf{x}_t)]^T$.

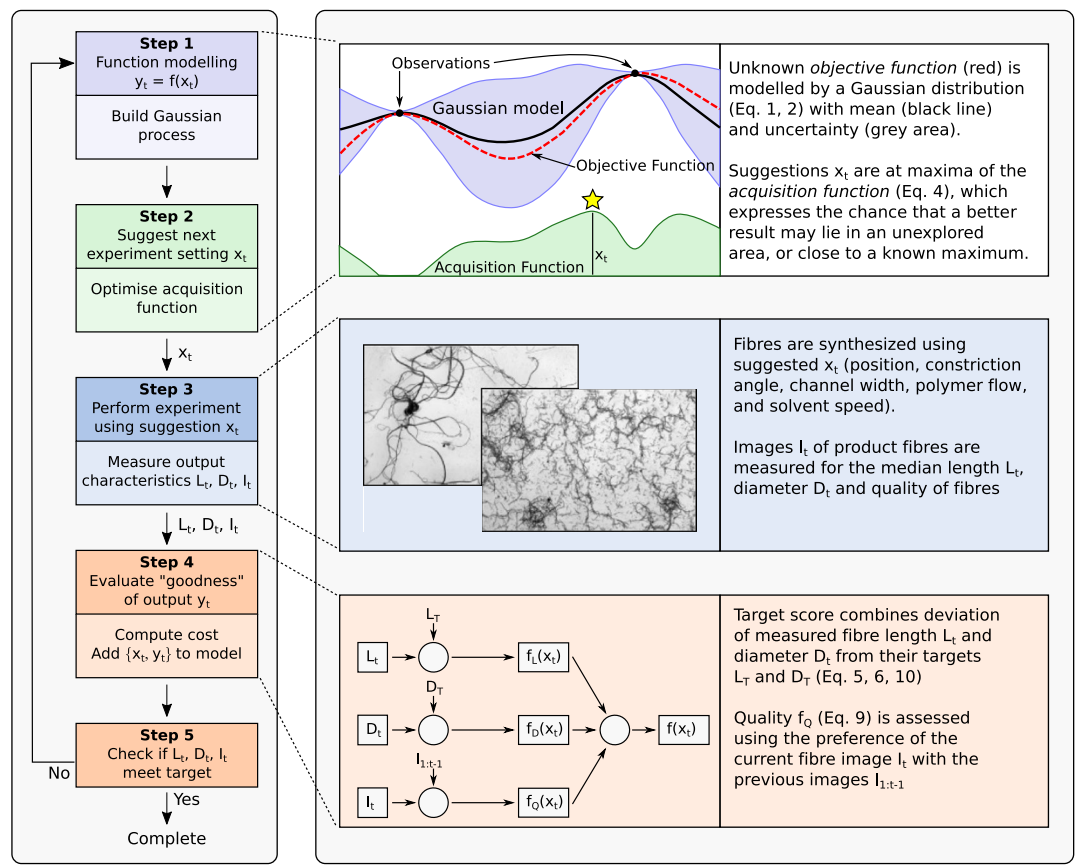

**Figure 2.** Adaptive Experimental Optimisation. Five steps include: (1) modelling the unknown function, (2) recommending the next experimental setting, (3) manufacturing and measuring short polymer fibers, (4) evaluating the product utility, and (5) checking target. These steps repeat until the target is achieved, or the process is terminated.

**Step 2: Recommend next experimental setting.** The function $f$ is expensive to sample due to the potential cost of experimental measurements, hence an alternate strategy is to build a surrogate function that is cheaper to compute and finds its maximum. These acquisition functions trade-off *exploitation* of the high predicted mean with *exploration* using high predicted variance. We use the expected improvement (EI) as the acquisition function, which defines the improvement over the current best value and is given as ref. [12]

$$a_{EI}(\mathbf{x}) = \begin{cases}(\mu(\mathbf{x}) - f(\mathbf{x}^+))\Phi(Z) + \sigma(\mathbf{x})\phi(Z) & \sigma(\mathbf{x}) > 0 \\ 0 & \sigma(\mathbf{x}) = 0\end{cases} \quad (4)$$

where $Z = \frac{\mu(\mathbf{x}) - f(\mathbf{x}^+)}{\sigma(\mathbf{x})}$. $\Phi(Z)$ and $\phi(Z)$ are the CDF and PDF of standard normal distribution.

To maximise the acquisition function, we use DIRECT[22], a deterministic and derivative-free optimiser. The resulting maximum is the recommendation for the next experimental setting.

**Step 4: Deriving the combined target score $y_t$.** *Quantitative Scores.* Consider $\mathbf{x}_t$ as the vector of experimental parameters (solvent flow rate, polymer flow rate, device angle, device position and channel width) at iteration $t$. We formulate quantitative scores, $f_L(\mathbf{x}_t)$ and $f_D(\mathbf{x}_t)$ as the difference of the measured fiber median length $L_t$ and fiber median diameter $D_t$ to their respective targets $L_T$ and $D_T$ as:

$$f_L(\mathbf{x}_t) = \frac{|min\{L_{max}, L_t\} - L_T|}{L_{max} - L_T} \quad (5)$$

$$f_D(\mathbf{x}_t) = \frac{|min\{D_{max}, D_t\} - D_T|}{D_{max} - D_T} \quad (6)$$

The maximum values $L_{max}$ and $D_{max}$ are chosen as target bounds beyond which the fibers are not useful. In our case, $L_{max} = 3L_{Target}$, and $D_{max} = 3D_{Target}$.

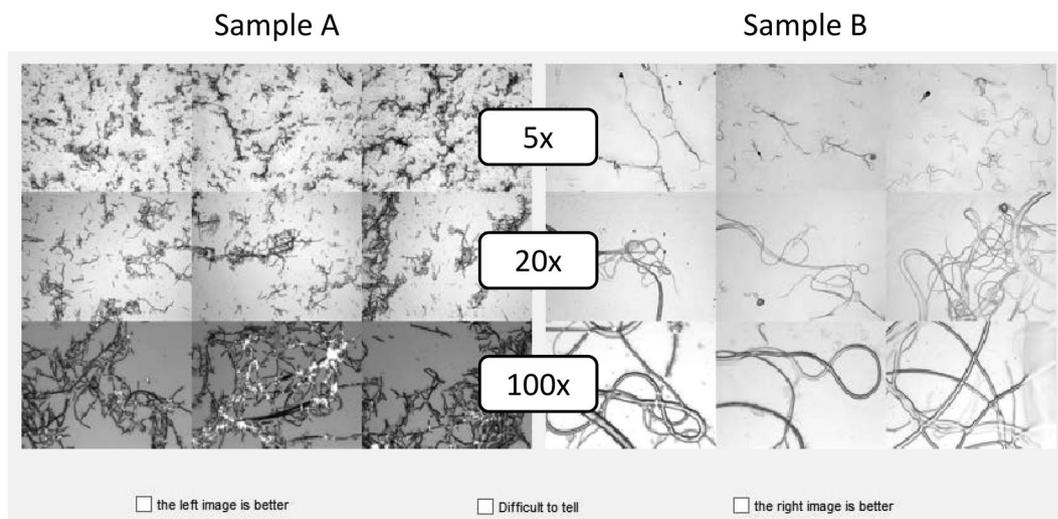

**Figure 3.** Fiber quality comparison example. The montage of fiber images from two experimental settings are compared visually. For each experiment there are three images for each of three different magnifications: 5x (top row), 20x (centre row) and 100x (bottom row). In this example, the "A" sample on the left is preferred because it is more homogeneous, even though some debris is present. The "B" sample is cleaner, but is not homogeneous, and contains some fibers that are too large.

*Qualitative Score - $f_Q(\cdot)$ and interface for qualitative comparison.* The qualitative score captures aspects that are difficult to measure quantitatively for example, uniformity and nature of the precipitate's morphology, amount of by-product, homogeneity etc. It is difficult to rank each fiber sample image on a pre-determined and uniformly-spaced scale which may cover all types and qualities of samples, since at any point of the experiment the iteratively-collected samples are not representative of the whole spectrum of possible samples and the scaling might change as a result and be prone to high subjective bias. Possible solutions include the preparation of a representative sample for each unit of the scale, but this would void the efforts for increased efficiency.

An alternative strategy is to use a visual interface to compare pairs of images, scoring them as either better, worse or difficult to tell. Since this categorisation is broad, subjective bias is reduced. Although it would be useful, it is not practical to rank the current image against all previous images. At any iteration $t$, the method asks the user to compare the images of the fibers produced in the current iteration ($I_t$ produced by parameter configuration $x_t$) with a set of images randomly chosen from previous iterations. If image $I_t$ generated from configuration $x_t$ is preferred over image $I_k$ from configuration $x_k$, we get $x_t \succ x_k$. If $I_k$ is preferred over $I_t$, we get $x_k \succ x_t$. If it is difficult to tell, both relationships are maintained: $x_t \succ x_k$, and $x_k \succ x_t$ which ensures that the qualitative scores returned by Equation (9) are close.

This process gives rise to a set of observed pairwise preferences at iteration $t$ that we denote as the preference set $\Omega_t$:

$$\Omega_t = \{x_i \succ x_j\} \qquad (7)$$

We formulate a method to extract qualitative scores from such pairwise comparisons. The number of comparisons scales logarithmically with iteration number $t$ as $\lfloor \log_2(t-1) \rfloor + 1$. The interface for image comparison is shown in Fig. 3. The user can indicate which image is better or if it is difficult to tell. The qualitative assessment of the samples favours high fiber yield, low entanglement (homogenous dispersion) and absence of by-products such as debris and spheres. In this example, the "A" sample on the left is preferred because it is more homogeneous, even though some debris is present. The "B" sample is cleaner, but is not homogeneous, and contains some very large fibers.

From this data, we learn the latent score function $f_Q$ at each iteration.

*Computation of $f_Q$.* Given a preference set $\Omega$ defined in Equation (7), we use a Gaussian process $p(f_{LQ}|\Omega)$ to learn the latent relation $f_{LQ} = [f_{LQ}(\mathbf{x}_1), f_{LQ}(\mathbf{x}_2), \cdots]^T$ that explains this partial ranking across the images at each iteration. As the set of images differ between iterations, the random choice at each iteration will produce different images to compare. The function explaining the latent ranking relationship will keep improving but will change from one iteration to the next.

The unobserved latent function $f_{LQ}(\mathbf{x}_i)$ is associated with each sample $\mathbf{x}_i$. We impose a Gaussian process prior on these latent values and derive the latent values based on an approximate likelihood function. We assume that the latent function is contaminated with Gaussian noise $\mathcal{N}(\delta; \mu, \sigma_{noise}^2)$.

Suppose that the preference set $\Omega$ contains $M$ preference pairs $\Omega = \{u_m \succ v_m, m = 1, \ldots, M\}$ and denote the $m$th preference pair $u_m \succ v_m$. The likelihood function is defined as ref. 23

$$p(u_m \succ v_m | f_{LQ}(u_m), f_{LQ}(v_m)) = \Phi(z_m)$$

where $z_m = \frac{f_{LQ}(u_m) - f_{LQ}(v_m)}{\sqrt{2}\sigma_{noise}}$ and $\Phi(z) = \int_{-\infty}^{z} \mathcal{N}(\gamma; 0, 1) d\gamma$.

We want to maximise the posterior distribution of the latent function, or

$$p(f_{LQ}|\Omega) \propto p(f_{LQ}) \prod_{m=1}^{M} p(u_m \succ v_m | f_{LQ}(u_m), f_{LQ}(v_m))$$

where $p(f_{LQ})$ is the prior probability and is a multi-variant Gaussian function with the covariance function $\Sigma$. Computing the maximum a posterior (MAP) estimate of the latent function is equivalent to minimising the following function

$$\mathcal{S}(f_{LQ}) = -\sum_{m=1}^{M} \ln \Phi(z_m) + \frac{1}{2} f_{LQ}^T \Sigma^{-1} f_{LQ}$$

We can employ the Newton-Raphson approach to find the solution for simple cases. By setting $\left.\frac{\partial \mathcal{S}(f_{LQ})}{\partial f_{LQ}}\right|_{\widehat{f_{LQ}}} = 0$, we get the MAP estimate of $f_{LQ}$ as

$$\widehat{f_{LQ}} = \Sigma \beta \qquad (8)$$

where $\beta = \left.\frac{\partial \sum_{m=1}^{M} \ln \Phi(z_m)}{\partial f_{LQ}}\right|_{\widehat{f_{LQ}}}$. Further details of this approach are described in the literature[23]. To avoid over-stretch or over-squeeze on quality scores, we use the following normalisation approach

$$f_Q = \frac{\widehat{f_{LQ}} - \min(0, \widehat{f_{LQ_{min}}})}{\max(\widehat{f_{LQ_{max}}} - \min(0, \widehat{f_{LQ_{min}}}), 1)} \qquad (9)$$

where $\widehat{f_{LQ_{max}}}$ and $\widehat{f_{LQ_{min}}}$ are the minimal and maximal value of $\widehat{f_{LQ}}$ respectively; and $f_Q$ are the normalised quality scores.

*Combining quantitative and quality scores.* We combine the multiple objectives into a single objective score. Since $f_L(\mathbf{x}_t)$ and $f_D(\mathbf{x}_t)$ are difference to the targets, the values need to be minimised while $f_Q(\mathbf{x}_t)$ needs to be maximised because higher values are indicative of better quality. The combined utility score is defined as proportional linear combination. In the absence of any prior knowledge, we assign equal weight (1/3) to length, diameter and quality.

$$y_t = f(\mathbf{x}_t) = \frac{1}{3} \times f_L(\mathbf{x}_t) + \frac{1}{3} \times f_D(\mathbf{x}_t) + \frac{1}{3} \times (1 - f_Q(\mathbf{x}_t)) \qquad (10)$$

The score $y_t$ has range [0, 1] and is 0 when all objectives are met. Our optimiser seeks to minimise this score, which increases with the deviation of product fibres from the target. This score will not decrease monotonically because the objectives may get better or worse individually across iterations. To track the objective function progress towards targets we define the *BestFoundValue*$_t$ (BFV) to be the lowest value of $y_t$ found until iteration $t$.

**Data Availability.** The data that support the findings of this study are available from the corresponding author upon reasonable request.

## Results

Our experiments aim to (1) find optimal conditions for manufacturing different fibres, and (2) validate the Adaptive Experimental Optimisation (AEO) method. Fiber production experiments were conducted iteratively, varying the five input variables (solvent flow rate, polymer flow rate, device angle, device position and channel width), until the product dimensions converged to within 20% of the target length and diameter. Initial trials found this to occur within about 20 iterations, but this number will vary with process and will rise as tolerance decreases.

We performed a series of experimental runs, each consisting of a total of 20 iterations guided by AEO, starting from five initial chosen synthetic conditions, which were either randomly-chosen or known-outcome conditions (random, bad, good). This process was repeated for three different fiber targets: (1) median length 70 μm and median diameter 1 μm, (2) length 40 μm and diameter 0.6 μm, and (3) length 50 μm and diameter 0.4 μm. For Target 1 we performed three runs with random starting states, and one run each from known "good" and "bad" starting states. For Targets 2 and 3 we performed two runs from random starting states. Table 1 summarises the results of optimisation experiments for three target fibers. The table lists the best found value (BFV), the iteration at which it was found, and the difference to target as a percentage of target length (L%) and diameter (D%). Most runs start from 5 randomly chosen states. We present here three typical results: target achieved from a random starting state, target achieved from a "bad" starting state, and target not achieved.

| Run # | Target | $L_T$ | $D_T$ | Start states | BFV | Iteration | L% | D% | Figure |
|---|---|---|---|---|---|---|---|---|---|
| 1 | 1 μm | 70 μm | 1 μm | Random | 0.221 | 14 | 0.5 | 4.8 | 4 |
| 2 | 1 μm | 70 μm | 1 μm | Random | 0.261 | 7 | 42.6 | 19.8 | |
| 3 | 1 μm | 70 μm | 1 μm | Random | 0.264 | 13 | 2.5 | 14.0 | |
| 4 | 1 μm | 70 μm | 1 μm | Bad | 0.161 | 12 | 0.5 | 4.8 | S.1 (Supplement) |
| 5 | 1 μm | 70 μm | 1 μm | Good | 0.204 | 3 | 20.6 | 3.4 | |
| 6 | 2 μm | 40 μm | 0.6 μm | Random | 0.237 | 13 | 15.8 | 0.5 | |
| 7 | 2 μm | 40 μm | 0.6 μm | Random | 0.223 | 6 | 33.6 | 0.2 | |
| 8 | 3 μm | 50 μm | 0.4 μm | Random | 0.307 | 4 | 2.7 | 104 | |
| 9 | 3 μm | 50 μm | 0.4 μm | Random | 0.331 | 2 | 2.7 | 104 | S.2 (Supplement) |

**Table 1.** Results of optimisation for 9 experiments over three different targets. Each run is identified by a run number, target length, and diameter. The table shows the starting state (random, bad, or good), best found value (BFV) of the objective function, the iteration this was achieved, and the corresponding deviations from target length and diameter.

Figure 4 shows results for Target 1, a fiber of median length 70 μm and median diameter 1 μm. The BFV is shown in part (a). Parts (b) and (c) show the corresponding difference to target length and diameter as a percentage of the target. Iteration steps without circles have values above the maximum scale for the axis. The solid line indicates BFV, and the images at samples where the BFV is improved are shown in (d). The lowest BFV is achieved at iteration 14, with a 4.8% distance from target in length (b) and 0.5% distance in diameter (c). Note that the objective function (shown by circles in (a)) does not decrease monotonically, since at each iteration one criterion may improve while others may worsen.

Further analysis for Target 1 (L = 70 μm, D = 1 μm) is considered by varying the starting conditions. Instead of starting from randomly chosen conditions, five states that are known (from preliminary "seed" experiments) to lead to a product which is very different from the specified target (Supplementary Fig. S.1) are used. The target set were reached within 12 iterations.

For Target 1 (L = 70 μm, D = 1 μm) runs labelled "Good" and "Bad" start from the five best and worst samples achieved in the previous three runs. Note that the "Bad" run finds the same solution as Run 1, which is within 0.5% for target length, and 4.8% for target diameter. The BFV for this result is lower than achieved for Run 1 (0.161 < 0.221) suggesting that a "bad" start may cause a subsequent increase in the quality ratings.

Table 2 summarises the differences between targets. Target 1 (L = 70 μm, D = 1 μm) and Target 2 (L = 40 μm, D = 0.6 μm) show similar values for mean BFV and iteration. However, the average L% is significantly higher for Target 2, with fewer experiments showing a low value for L%, indicating that the process produces a large portion of fibers of length above 100 μm. Target 3 (L = 50 μm, D = 0.4 μm) shows a higher (worse) BFV of 0.32, and none of the tested conditions achieved the target diameter of 0.4 μm. Both runs find the same solution (L% = 2.7, D% = 104) within 4 iterations, but no subsequent improvement was obtained in the next 16 iterations. It should be noted that this finding is to be expected in some scenarios because defining a target does not imply that the system is capable of achieving it.

## Discussion

The AEO process described has allowed fast optimisation of a complex and multi-variable "black-box" system, where short polymer fibers of specified target diameter, length and quality could be produced using bespoke devices based on a small pool of preliminary data. The necessity to combine heterogeneous criteria is very common in materials synthesis process optimisation, as are high-degrees of complexity and low predictability of systems studied. The SPF production process here described involves a complex interplay of forces to result in the product, and a variety of different material properties can be produced as a result. The SPF process is dominated by local fluid-dynamics and the thermodynamics of polymer dope solidification. As the filament is flowing through the dispersant, it undergoes both elongation and coagulation. The forming short fibres therefore exhibit continuously-varying viscosity (which also displays a radial gradient within the filament) and experience a continuously-varying flow field (external forces). Modifying control parameters such as flow rates, therefore, has a double effect: on the fluid-dynamics and on the viscosity of the filament as function of position and time. Additional complications in attempting to build an accurate micro-mechanical model include: the low-level accuracy in the local flow-field knowledge, the non-constant-flow output by the pumps (which is pulsating), and the effects of device roughness. This system is, therefore, best optimised through output analysis, but the large matrix of conditions cause this to be a very time-consuming challenge. Its optimisation through iterative processes guided by AEO was here demonstrated possible. We set about controlling not just the quality and uniformity of the produced sample morphology, but also the sample's median dimensions, in a system which could suffer from non-orthogonality in the process parameter-outcome relationship against the different criteria. The process has allowed us to combine different types of information: quantitative (scalar) and qualitative (categorical, non-uniform) and to optimise the combination of the three.

The AEO process will never self terminate as the iterative process loop will continually search for optimal states. Thus we can never say with certainty that a particular experimental setup is not capable of reaching a target. However, tracking the BFV can provide useful insight. If BFV does not decrease further with increasing iterations, it is likely but not certain the experimental system will not meet the chosen target. Supplementary an informed manner, and will be able to apply judgement to a non-improving BFV, where also taking into account cost of experimentation vs likelihood of results.

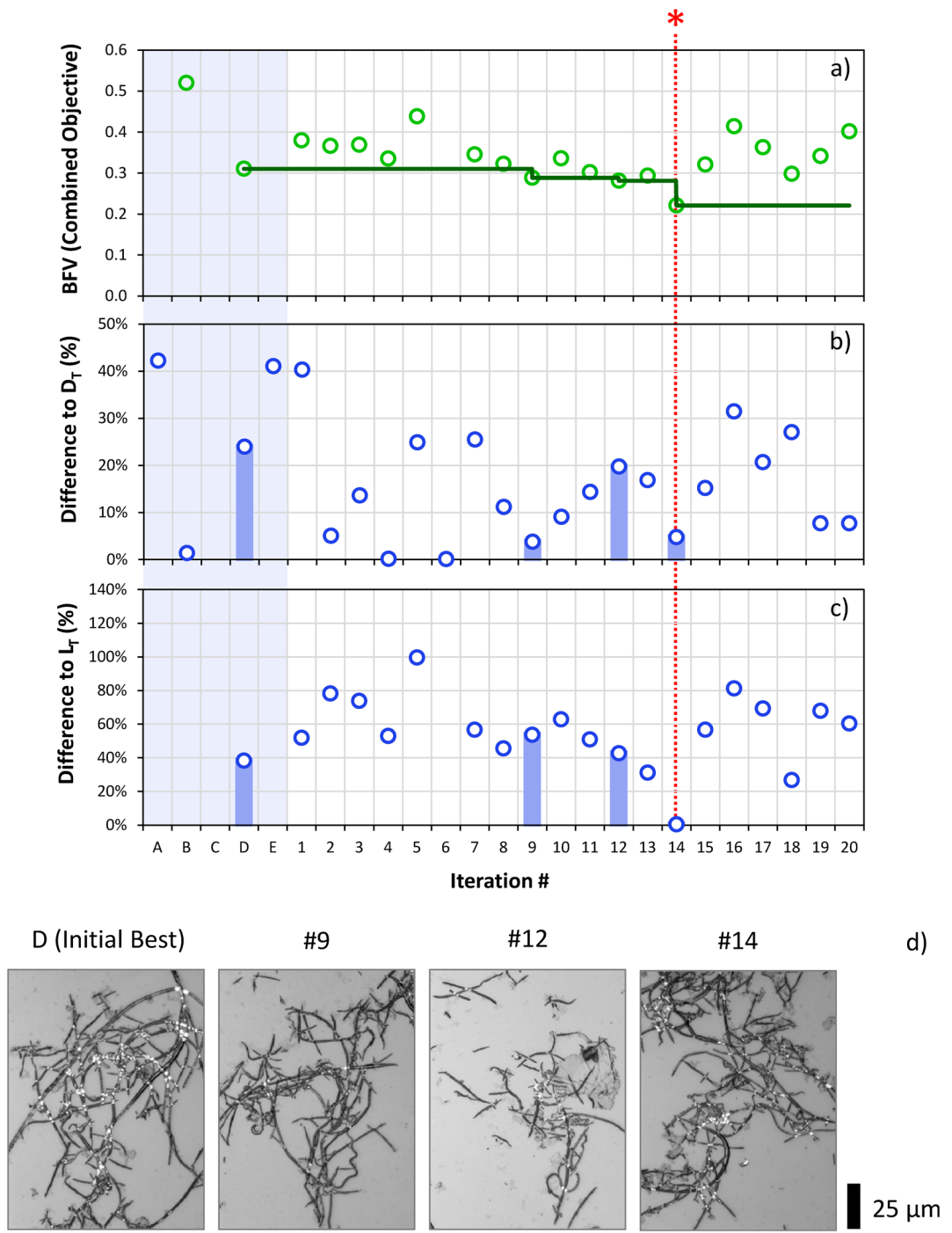

**Figure 4.** The results of a Run 1, starting from random samples for Target 1 (L = 70 μm, D = 1 μm). (**a**) shows the Best Found Value (BFV) of the combined objective for iterations of the optimisation process. The vertical axis is the dimensionless combined objective, in the range 0 to 1. Lower values are better. The horizontal axis shows iteration number; Samples A–E represent the 5 initial random samples. The value at each iteration is indicated by a circle. (**b**) $|L - L_T|/L_T$, the magnitude of the deviation in found length $L$ from the target $L_T$, as a percentage of the target $L_T$. (**c**) $|D - D_T|/D_T$, the magnitude of the deviation in found diameter $D$ from the target $D_T$, as a percentage of the target $D_T$. These two figures show how closely the BFV is satisfying the specified target criteria. The vertical bars correspond to the deviations in length and diameter at point where the BFV improves. (**d**) The corresponding SPF optical microscope images. The red asterisk (*) and dashed line indicates the optimal settings identified in the iterations performed.

Figure S.2 shows the results for an unachievable target: L = 50 μm, D = 0.4 μm. From iterations 2–20, the BFV does not decrease. Though the length target was found to be achievable, the deviation from the diameter target remained poor (>100% from target). In a real setting, it is likely that an experimenter will approach the results in

| Target | $L_T$ | $D_T$ | mean BFV | mean Iteration (range) | mean L% (range) | mean D% (range) |
|---|---|---|---|---|---|---|
| 1 | 70 μm | 1 μm | 0.25 | 9.8 (3–14) | 13.3 (0.5–42.6) | 9.4 (3.4–19.8) |
| 2 | 40 μm | 0.6 μm | 0.23 | 9.5 (6–13) | 24.5 (15.8–33.6) | 0.35 (0.2–0.5) |
| 3 | 50 μm | 0.4 μm | 0.32 | 3 (2–4) | 2.7 (2.7) | 104 (104) |

**Table 2.** Summary of optimisation by fiber target.

In order to design and operate the process it is important to know which variables have the strongest influence on performance. To estimate this we compared each state with the length and diameter of the resulting fibers, performing second order polynomial fit using samples over all 9 experiments, and noting the value of the resulting correlation coefficients R. The correlations between process parameters and the Length and Diameter is contained in Supplementary Table S.1. The angle and position have very little influence on the characteristics of the produced fibers. Polymer flow has a moderate influence (0.37) on length, but only a weak influence on diameter. Thus the most significant influence on overall performance is solvent speed followed by channel width. Sensitivity analysis shows that solvent speed also accounts for the majority (34%) of the uncertainty. See Supplementary Table S.2 for more details. While it is to be noted that these results are only valid within the set of tested experiment states, they provide an initial useful set of indications on the system as a whole.

While offering significant advantages for the optimisation of material properties and process configuration for SPF examples, there remains significant scope to expand the capability of experimental optimisation through AEO to other experimental systems. New methods must be found to increase the number of input variables and output target variables. Current algorithms finding systems with more than just 10 control variables is a challenge[24, 25]. The calculation of Expected Hypervolume Improvement, a key step in multi-objective optimisation quickly becomes computationally unfeasible as the number of objectives rises[26]. These problems present open computational challenges.

## Concluding Remarks

We have formulated and demonstrated Adaptive Experimental Optimisation (AEO) incorporating both material properties and process characteristics within a coherent experimental optimisation framework. Heterogeneous and multiple objectives can be successfully combined for experimental optimisation with the method. We show how to incorporate both qualitative and quantitative targets in an unified experimental optimisation setting. An efficient methodology to systematically elicit qualitative information about the experimental merit is presented. Through the AEO methodology, we have demonstrated how a developmental experimental system such as SPF synthesis can be efficiently directed to achieve material and process objectives. We show the potential for using the AEO methodology as a new kind of intelligent agent where the experimenter and the agent explore the new space together providing a mechanism for yielding greater insight into the relationships between critical variables, and a path to more efficient optimisation of novel materials and processes.

### Acknowledgements

This research was supported under Australian Research Council's Industrial Transformation Research Hub funding scheme (project number IH140100018). The present work was carried out with the support of the Deakin Advanced Characterisation Facility. The authors are thankful to Mr Keiran Pringle for assisting with device manufacture.

### Author Contributions

A.S. directed design and operation of the SPF manufacturing at IFM. T.S. assisted with experiments. D.R.d.C.L. ran trials and prepared results. S.V. directed design of the AEO experimental optimiser at PRaDA. S.R. and S. Gupta designed the optimiser. C.L. ran field trials and prepared results. S. Greenhill, S.V., A.S. and M.H. interpreted results and prepared the manuscript. M.H. also acted as industry advisor. All authors discussed the results and revised the manuscript.

### Additional Information

**Supplementary information** accompanies this paper at doi:10.1038/s41598-017-05723-0

**Competing Interests:** The authors declare that they have no competing interests.

**Publisher's note:** Springer Nature remains neutral with regard to jurisdictional claims in published maps and institutional affiliations.